\documentclass[letterpaper, 10 pt, conference]{ieeeconf}  

\IEEEoverridecommandlockouts                              
\usepackage{cite}
\usepackage{amsmath,amssymb,amsfonts}
\usepackage{algorithmic}
\usepackage{graphicx}
\usepackage{textcomp}
\usepackage{xcolor}
\usepackage{multirow}
\usepackage{booktabs}
\usepackage{url}
\usepackage{subfigure}

\usepackage{times} 
\usepackage[colorlinks,
            linkcolor=black,
            anchorcolor=black,
            citecolor=black]{hyperref}

\title{\LARGE \bf
Pose Refinement with Joint Optimization of Visual Points and Lines
}

\author{Shuang Gao$^{1,\dagger}$, Jixiang Wan$^{1,2,\dagger}$, Yishan Ping$^{1,\dagger}$, Xudong Zhang$^{1,\ast}$, Shuzhou Dong$^{1}$, \\
Yuchen Yang$^{1}$, Haikuan Ning$^{1}$, Jijunnan Li$^{1}$, Yandong Guo$^{1}$
\thanks{$^{1}$ All authors are with OPPO Research Institute, Shanghai, China.}
\thanks{$^{2}$ is also with Department of Automation, Shanghai Jiao Tong University, Shanghai, China.}
\thanks{${\dagger}$ Authors contributed equally to this work.}
\thanks{${\ast}$ indicates corresponding author. Contact: zhangxudong@oppo.com}
}

\begin{document}

\maketitle
\thispagestyle{empty}
\pagestyle{empty}

\begin{abstract}

	High-precision camera re-localization technology in a pre-established 3D environment map is the basis for many tasks, such as Augmented Reality, Robotics and Autonomous Driving. The point-based visual re-localization approaches are well-developed in recent decades, but are insufficient in some feature-less cases. In this paper, we design a complete pipeline for camera pose refinement with points and lines, which contains the innovatively designed line extracting CNN named VLSE, the line matching and the pose optimization approaches. We adopt a novel line representation and customize a hybrid convolution block based on the Stacked Hourglass network \cite{newell2016stacked}, to detect accurate and stable line features on images. Then we apply a geometric-based strategy to obtain precise 2D-3D line correspondences using epipolar constraint and reprojection filtering. A following point-line joint cost function is constructed to optimize the camera pose with the initial coarse pose from the pure point-based localization. Sufficient experiments are conducted on open datasets, i.e, line extractor on Wireframe and YorkUrban, localization performance on InLoc \textit{duc1} and \textit{duc2}, to confirm the effectiveness of our point-line joint pose optimization method.

\end{abstract}

\section{INTRODUCTION}

Re-localization technology based on visual features is widely used in the field of Computer Vision, such as Augmented Reality, Indoor Navigation, Mobile Robots and Autonomous Driving\cite{lim2012real}\cite{castle2008video}. The essence of visual localization is to obtain the camera's 6 degrees of freedom (DoF) including its position and orientation through the corresponding relationship between the captured image and the pre-established 3D environment map. Common visual features on images include points, lines, planes and some more advanced features, like semantic labels.

The visual localization based on point features thrives in recent years. A large number of visual point extractors \cite{detone2018superpoint}\cite{revaud2019r2d2}, matching methods\cite{sarlin2020superglue}\cite{DBLP:journals/corr/abs-2104-00680} and point-based 3D mapping approaches \cite{schonberger2016structure}\cite{moulon2016openmvg} have been proposed. Although the visual points can lead to more accurate image positions, they are still sensitive to weak textures and illumination changes. In some cases, visual lines, containing more structured information, are more robust to these phenomenons.

For visual line detection, traditional methods such as LSD \cite{von2008lsd}, FLD \cite{lee2014outdoor} and EDlines \cite{akinlar2011edlines} usually rely on low-level image gradients and edge features. Moreover, they also struggle to distinguish high-level local textures and structural elements. 
The Convolutional Neural Networks (CNN) based line segment detection models have achieved superior performances over the traditional methods \cite{huang2018learning, zhou2019end, zhang2019ppgnet, xue2019learning, xue2020holistically, huang2020tp, dai2021fully,zhang2021elsd, gu2021towards}. Supported by a large-scale dataset with manual wireframe annotations \cite{huang2018learning}, we succeed in training a deep model to capture structured and informative line segments (e.g., building contours, stairs), while ignoring line segments associated with texture (e.g., carpet patterns), which aims to lift the subsequent mapping and localization accuracy.


In this work, we also present a novel pose optimization method, that combines the advantages of both point and line in the localization task. The major idea of the point-line joint optimization method is to adjust poses by minimizing the joint reprojection error of point and line correspondences. In detail, a standard pose estimation method based on point correspondences is employed to obtain an initial pose, and the inlier point correspondences are saved for subsequent optimization. The line segments on each image are extracted in both the mapping and localization stages. We innovatively design a CNN architecture called VLSE to extract the visual lines. And the line matching correspondences will be acquired based on the initial pose and 3D line map. Finally, a point-line joint optimization cost function is employed to improve the accuracy of the final pose estimation.

In this paper, we conclude our contributions 
mainly into the following parts:




1. We design an integrated pipeline for camera pose refinement in visual localization with joint optimization of points and lines. It has great flexibility and each module can be replaced and promoted individually.

2. We propose a more accurate visual line detection CNN named VLSE, where we utilize an innovative line segment representation and a customized hybrid convolution block based on the Stacked Hourglass network\cite{newell2016stacked}.

3. We design a more robust and accurate cost function combining point and line for line matching and pose optimization. And we adopt a pure geometric-based line matching method to avoid the difficulty of visual-based matching caused by changes in illumination, scale and perspective.

\begin{figure*}[htbp]
	\centering
	\includegraphics[width=0.9\textwidth]{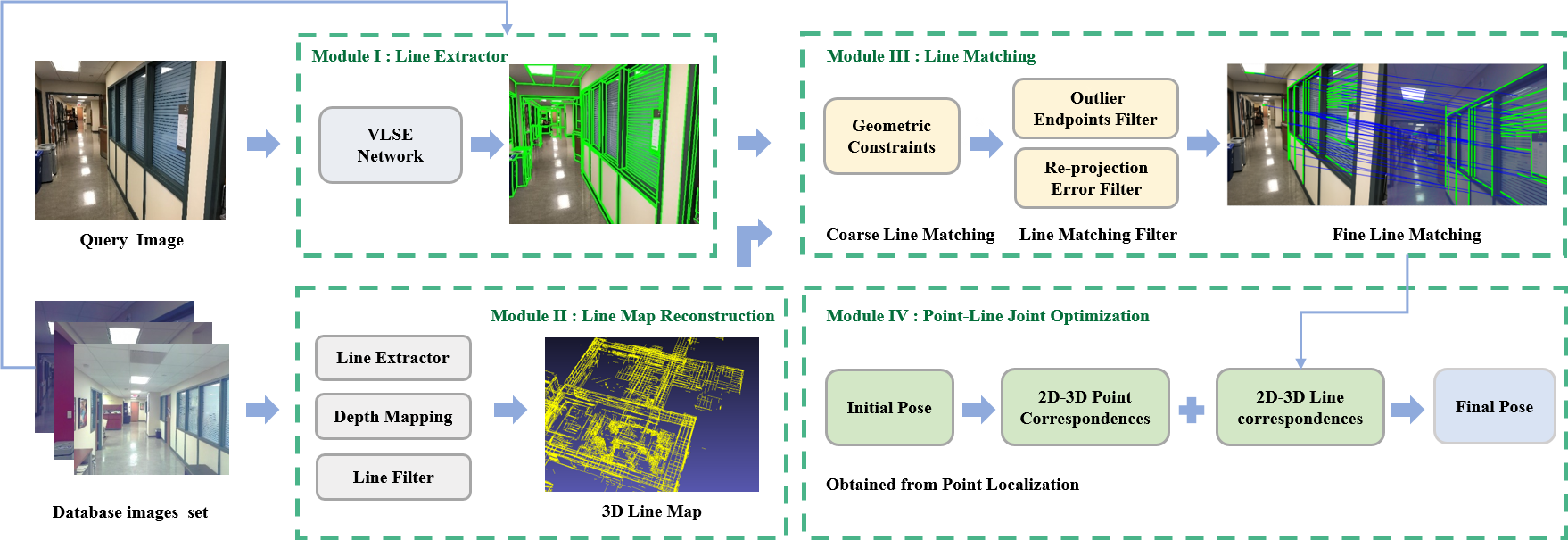}
	\caption{
		\textbf{Point-Line Joint Pose Optimization Pipeline}. The input data contains database images set for mapping, query image with initial pose and 2D-3D point correspondences from the initial pure point-based localization. Module I is visual line extraction for subsequent line matching and mapping. The output of Module II is a 3D line map reconstructed. Module III is a geometric-based line matching and its input contains 2D lines from Module I, a 3D line map from Module II and the initial pose of the query image. Module IV is point-line joint optimization.}
	\label{fig:pipeline}
\end{figure*}

\section{RELATED WORKS}

In this section, we introduce some recent progress in visual line extraction, line matching and line-based visual localization researches. All of them are highly related to our work on point-line optimization on the pose refinement task.

\textbf{Deep Line Segment Detection.} \cite{huang2018learning} provides a large public dataset with wireframe annotations and a heuristic wireframe fusion algorithm, which makes the learning-based line segment detection task possible. Moreover, \cite{zhou2019end} and \cite{zhang2019ppgnet} train end-to-end neural networks to filter the wrong predicted endpoints in the junction heatmap. \cite{xue2020holistically} integrates AFM-based reparameterization scheme \cite{xue2019learning} for line segments, and significantly improves both accuracy and efficiency. Recently, \cite{huang2020tp, dai2021fully,zhang2021elsd, gu2021towards} directly encodes a line segment into several parameters predicted directly from the networks, without heuristics-driven line proposal generation, and further simplify the architecture.

\textbf{Line Matching.} The line matching methods can be roughly divided into two categories. 
For traditional methods, \cite{hartley1995linear} uses the innovatively designed tensor to describe the lines and match them between images. \cite{zhang2013efficient} proposes the line band descriptor based on line appearance similarity, and the K Nearest Neighbors (KNN) match method can be used for matching. For learning-based methods, \cite{ma2020robust} uses a graph convolution network to match line segments. \cite{pautrat2021sold2} proposes a self-supervised occlusion-aware line description and detection method. Many methods face great difficulties in precision, especially when there is great viewpoint change or large-scale change between the matching image pairs.

\textbf{Line Localization.} For visual localization problems based on the line features, there are several works combining point features and line features in recent years.\cite{smith2006real} describes how straight lines can be added to a monocular Extended Kalman Filter Simultaneous Mapping and Localization (SLAM) system to provide good poses and mapping. \cite{gomez2019pl, pumarola2017pl, wei2019real} are PL-SLAM systems that defined distances from endpoints of 3D line to corresponding 2D line as line reprojection error. 
In \cite{humenberger2020robust}, the camera poses and 2D-3D line correspondences are iteratively optimized by minimizing the projection error of correspondences and rejecting outliers, where the line reprojection error denotes the sum of distances from endpoints to the corresponding 2D line. 

\section{Proposed Approach}

Our pose refinement pipeline mainly consists of four parts: 2D line extractor, 3D line map reconstruction, line matching and filtering, point-line joint optimization, which is illustrated in Fig. \ref{fig:pipeline}.

\textbf{Line extractor.} For line feature re-localization, an accurate and stable line extractor is needed. We design a CNN named as VLSE, described in \ref{lines extractor} to fulfill the extraction tasks both on mapping and query stages.



\textbf{Line map reconstruction}. For the mapping datasets with known global coordinates, we adopt a light-weight reconstruction method. We map the 2D line to the global coordinate directly, and a PnP with RANSAC is performed to filter the 3D outliers. The global coordinates are from depth sensors or dense reconstruction.

\textbf{Line matching and filtering.} The candidate 2D line matches between database image and query image are calculated by utilizing epipolar constraint based on the initial pose of query obtained from point localization. Then we map the 2D line on database images onto the 3D line map, and deny the line matches with a large reprojection error.

\textbf{Point-Line joint optimization.} Given the initial pose and corresponding points matches from point-based localization, we implement a process to minimize reprojection error of point correspondences and line correspondences to make the final estimated pose more accurate.

The following contents are constructed as our three main contributions in the total pipeline: visual line segment extractor, line matching and filtering during 3D line mapping, and pose refinement with point-line.


\subsection{VISUAL LINE SEGMENT EXTRACTOR}
\label{lines extractor}

\subsubsection{\textbf{Line Representation}}
\label{line representation}

\begin{figure}
	\hspace{0.2cm}
	\subfigure[]{
		\hspace{0.9cm}
		\begin{minipage}[b]{0.2\textwidth}
			\includegraphics[width=0.7\textwidth]{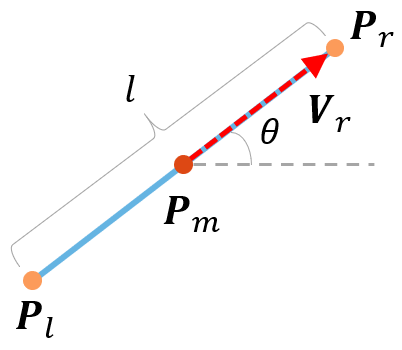}
		\end{minipage}
		\label{fig:Line_Rep_a}
	}
	\hspace{-15mm}
	\subfigure[]{
		\hspace{0.9cm}
		\begin{minipage}[b]{0.2\textwidth}
			\includegraphics[width=0.7\textwidth]{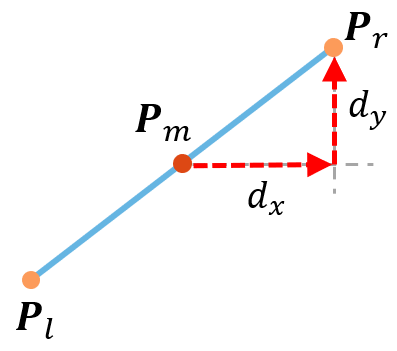}
		\end{minipage}
		\label{fig:Line_Rep_b}
	}
	\caption{ \textbf{Line segment representation.} For a line segment, one of the endpoints $\textbf{\textit{P}}_r$ can be calculated with the absolute location of the midpoint $\textbf{\textit{P}}_m$ and its relative position relationship to the midpoint $\textbf{\textit{V}}_r$, the other endpoint can also be determined by $\textbf{\textit{V}}_r$ in the opposite direction, due to the central symmetry of the two endpoints. (a) shows the traditional line representation with length $l$ and angle $\theta$. (b) shows the proposed line representation with horizontal and vertical distance  $d_x$ and $d_y$ from right point to the midpoint.}
	\label{fig:Line_representation}
\end{figure}

Inspired by \cite{dai2021fully, zhang2021elsd}, instead of proposing complicated line candidates by using the line-sampler in L-CNN \cite{zhou2019end}, we directly predict the locations of the two endpoints representing an effective line segment, which can save a great deal of memory and time consumption during post-processing. As depicted in Fig. \ref{fig:Line_representation}, $\textbf{\textit{P}}_l$ and $\textbf{\textit{P}}_r$ are the left and right endpoints of a line, while $\textbf{\textit{V}}_r$ is the vector representing the relationship between the $\textbf{\textit{P}}_r$ and the midpoint $\textbf{\textit{P}}_m$. Usually, the length $l$ and horizontal angle $\theta$ are used to produce the $\textbf{\textit{V}}_r$ \cite{dai2021fully, zhang2021elsd} as shown in Fig. \ref{fig:Line_Rep_a}. However,  the inaccuracy of angle $\theta$ will bring large line errors, especially for the prediction of a long line segment, because even a slight deviation of $\theta$ can have a great influence on the location of the endpoints. Such problems are relieved after we change to predict the pair of the horizontal distance $d_x=\frac{1}{2}(l\;cos\;\theta)$, because the junctions' position offset distance errors are more discriminative than the angle error during the training stage. Furthermore, $d_x$ and $d_y$ are adopted to produce $\textbf{\textit{V}}_r$ instead of the pair of length $l$ and $\theta$, as shown in Fig. \ref{fig:Line_Rep_b}. $\textbf{\textit{P}}_l$ and $\textbf{\textit{P}}_r$ are expressed by:
\begin{equation}
	\begin{split}
		\textbf{\textit{P}}_l=\textbf{\textit{P}}_m-\textbf{\textit{V}}_r, \quad
		\textbf{\textit{P}}_r=\textbf{\textit{P}}_m+\textbf{\textit{V}}_r,
	\end{split}
	\label{eq:Line_eq}
\end{equation}


\noindent where $\textbf{\textit{P}}_l$, $\textbf{\textit{P}}_m$ and $\textbf{\textit{P}}_r$ denote the coordinates of the left point, middle point and the right point of a 2D line, respectively.

\subsubsection{\textbf{Line Segment Detection}}

\begin{figure}[htbp]
	\centering
	\includegraphics[width=0.45\textwidth]{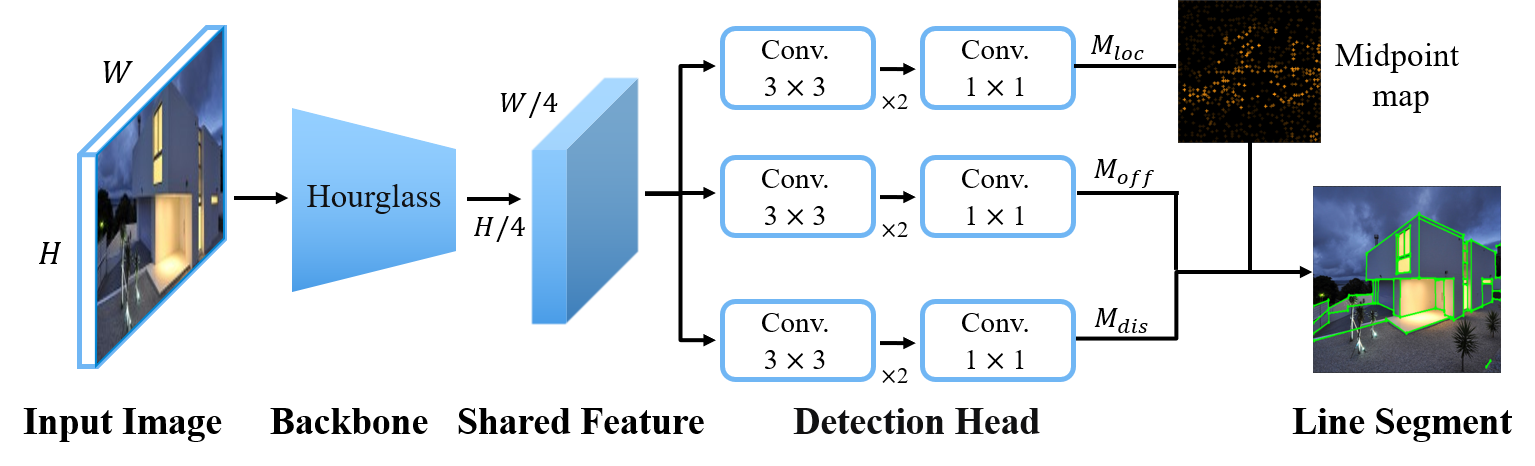}
	\caption{ \textbf{ The architecture of VLSE network}.  Stacked Hourglass network (HG) \cite{newell2016stacked} is chosen as our backbone to generate shared features. The following Convs are used to produce down-sampling and up-sampling features to form the midpoint map, and the line segment detection heads are used to construct the lines eventually.}
	\label{fig:VLSE_archi}
\end{figure}

\begin{figure}
	\hspace{-0.15cm}
	\subfigure[]{
		\hspace{0.1cm}
		\begin{minipage}[b]{0.2\textwidth}
			\centering
			\includegraphics[width=0.5\textwidth]{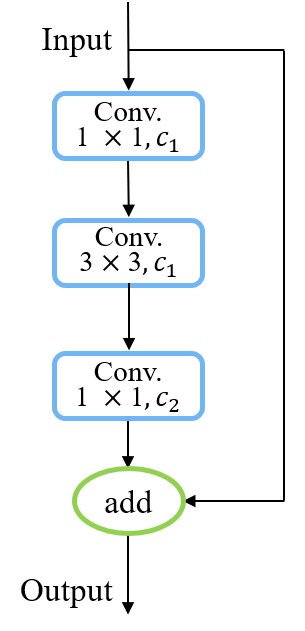}
		\end{minipage}
		\label{fig:Line_Block_a}
	}
	\subfigure[]{
		\hspace{-1.0cm}
		\begin{minipage}[b]{0.2\textwidth}
			\includegraphics[width=1.0\textwidth]{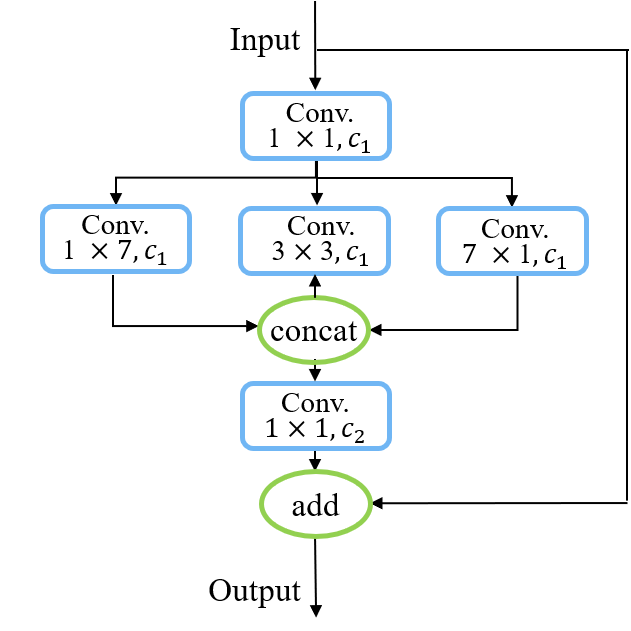}

		\end{minipage}
		\label{fig:Line_Block_b}
	}
	\caption{\textbf{Convolution blocks.} (a) is the basic convolution block in HG and (b) is our customized hybrid convolution block.}
	\label{fig:Line_Block}
\end{figure}

Fig. \ref{fig:VLSE_archi} illustrates our proposed VLSE (Visual Line Segment Extractor) architecture.
Which contains a feature extraction backbone and a line segment detection head. Stacked Hourglass network (HG) \cite{newell2016stacked} is chosen as our backbone to generate shared features, which is also adopted by L-CNN \cite{zhou2019end} and HAWP \cite{xue2020holistically}.
In which, the feature information at different scales is fused after a series of down-sampling and up-sampling. The basic convolution block of HG is shown in Fig. \ref{fig:Line_Block_a}, including two $1 \times 1$ convolution layers and one $3 \times 3$ convolution layer. Different from many other cases in Object Detection tasks, the length of the support region along the line's direction vector is much larger than the width of the region along the line's normal vector. Therefore, we additionally design two slender and long convolution kernels (e.g., $7 \times 1$ and $1 \times 7$). Drawing on the idea of Inception \cite{szegedy2015going}, we use the hybrid block concatenates the computational results from three different kinds of convolution kernels, instead of the original $3 \times 3$ convolution (as illustrated in Fig. \ref{fig:Line_Block_b}), as a way to fuse the visual features extracted from different receptive fields. After the backbone, the shared feature map, denoted by $\mathcal{F}$, is produced in the size of $H/4 \times W/4 \times 128$ with respect to the input image in the size of $H \times W \times 3$.

By feeding shared feature map $\mathcal{F}$ into the line segment detection head, we can further compute line segment proposals by predicting the midpoints and the locations of corresponding endpoints. Three feature maps are available through their respective detection head, as shown in Fig. \ref{fig:VLSE_archi}. Let $M_{loc}$ be the feature map representing the likelihood map, whose shape is $H/4 \times W/4 \times 2$, indicating whether each patch (the original image grid corresponded to each voxel on the feature map) contains the midpoint of a line segment or not. Then the midpoint confidence map can be calculated followed by a Softmax layer. In order to restore the pixel-level midpoint prediction in the original image, an additional feature map $M_{off}$ with the shape of $H/4 \times W/4 \times 2$ is also introduced to form the local offsets of the midpoints, recording the $x$ and $y$ displacement offset from top-left corner of patch, just like the junction predictions as intended in \cite{zhou2019end, xue2020holistically}. We adopt a sigmoid activation to normalize $M_{off}$ to the boundary with $[0,1) \times [0,1)$. The last feature map $M_{dis}$ represents the relative distance, which can infer the endpoint displacements with respect to the midpoints in the horizontal and vertical directions.

\subsection{LINE MATCHING AND FILTERING}
\label{lines matchimg section}

Inspired by Line3Dpp \cite{hofer2017efficient}, we propose a pure geometric-based matching method, which consists of acquiring 2D-2D candidate line match pairs by geometric constraint, mapping directly to 3D space with outliers removing, and selecting the most accurate 2D-3D line match pair by minimizing reprojection error defined in Eq. \ref{eq: line_costfuction}. The projection matrix of the query image is determined by the intrinsic and the initial pose. The initial pose is obtained from point-based localization. 

\begin{figure}[htbp]
	\centering
	\includegraphics[width=0.4\textwidth]{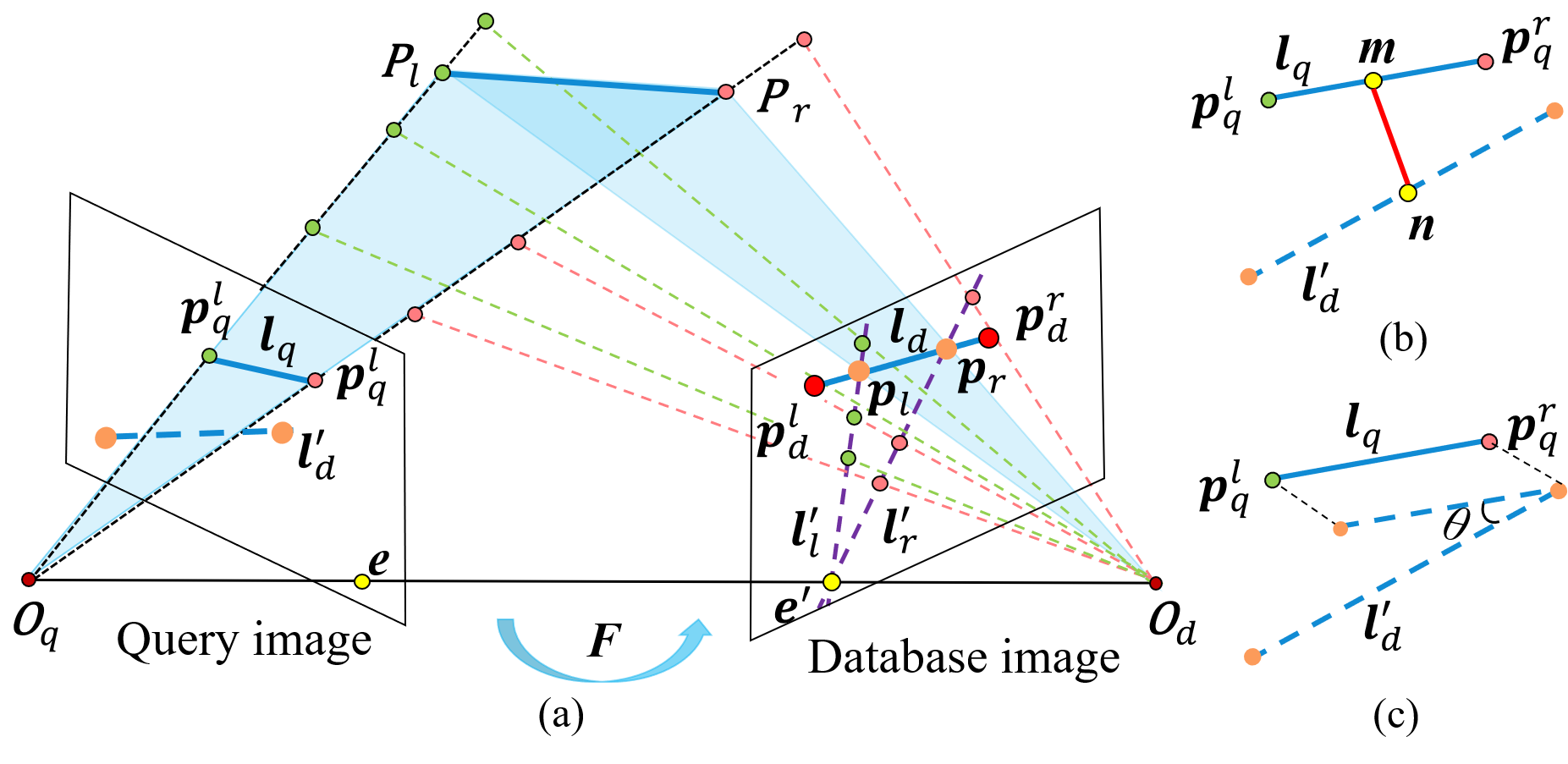}
	\caption{\textbf{The process of line matching.} (a) is the process of coarse line matching using geometry and depth reprojection constraints. (b) and (c) are the method to acquire high-quality 2D-2D line matches. $\textbf{\textit{l}}_{d}^{'}$ is projected line through world line $\textbf{\textit{P}}_l\textbf{\textit{P}}_r$ in the query image, point \textbf{\textit{m}} and \textbf{\textit{n}} represent the midpoint of line $\textbf{\textit{l}}_q$ and $\textbf{\textit{l}}_{d}^{'}$ respectively, $d$ represents the distance between point \textbf{\textit{m}} and point \textbf{\textit{n}}, $\theta$ is defined as the angle between line $\textbf{\textit{l}}_q$ and line $\textbf{\textit{l}}_{d}^{'}$.}
	\label{fig:line_matching}
\end{figure}

With the help of image retrieval \cite{zhou2021retrieval}, the image corresponding pairs between query images and database images are acquired, and line segments are extracted by VLSE described in Section \ref{lines extractor}. Inspired by \cite{andrew2001multiple} \cite{hofer2017efficient}, the first step of our matching algorithm is  employing epipolar constraints to obtain candidate line matches, which can be formulated by following:
\begin{equation}
	\begin{split}
		\textbf{\textit{l}}_{l}^{'}=\textbf{\textit{F}}*\textbf{\textit{p}}_{q}^{l}, \quad
		\textbf{\textit{l}}_{r}^{'}=\textbf{\textit{F}}*\textbf{\textit{p}}_{q}^{r},
	\end{split}
	\label{eq:Line_geometry_eq1}
\end{equation}
\begin{equation}
	\begin{split}
		\textbf{\textit{l}}_d= \textbf{\textit{p}}_{d}^{l} \times \textbf{\textit{p}}_{d}^{r},
	\end{split}
	\label{eq:Line_geometry_eq2}
\end{equation}
\begin{equation}
	\begin{split}
		\textbf{\textit{p}}_l= \textbf{\textit{l}}_d \times \textbf{\textit{l}}_{l}^{'}, \quad
		\textbf{\textit{p}}_r= \textbf{\textit{l}}_d \times \textbf{\textit{l}}_{r}^{'},
	\end{split}
	\label{eq:Line_geometry_eq3}
\end{equation}
\noindent where \textbf{\textit{F}} represents the fundamental matrix produced by the initial coarse pose of query and ground truth pose of the database image. As is illustrated in Fig. \ref{fig:line_matching}, $\textbf{\textit{l}}_q$ and $\textbf{\textit{l}}_d$ represent the line segments in the query image and the database image respectively. The $\textbf{\textit{p}}_{m}^{n}$ $(m = q,d$ and $n= l,r)$ represent homogeneous coordinates of the endpoints on the line segment. $\textbf{\textit{l}}_{l}^{'}$ and $ \textbf{\textit{l}}_{r}^{'} $ represent the epipolar line corresponding to point $\textbf{\textit{p}}_{q}^{l}$ and $\textbf{\textit{p}}_{q}^{r}$ respectively. $\textbf{\textit{p}}_l$ represents the intersect point of $\textbf{\textit{l}}_d$ and $\textbf{\textit{l}}_{l}^{'}$, $\textbf{\textit{p}}_r$ represents the intersect point of $\textbf{\textit{l}}_d$ and $\textbf{\textit{l}}_{r}^{'}$. The symbol * means matrix multiplication and the symbol $\times$ means vector cross product. The overlapping ratio between the matching lines will be computed\cite{schmid2000geometry}. And they will be retained as candidate line matches when their overlapping reached a certain threshold.

After the candidate 2D-2D line matches are obtained, we utilize the additional 3D line maps to establish 2D-3D matches. 
For mapping datasets, we first get the global coordinate of each pixel, which is from depth sensors or reconstruction. Then we adopt a light-weight reconstruction method to maps 2D lines to 3D space directly. For 3D endpoint coordinates with large noise, a PnP with RANSAC method is applied on database images to filter lines whose endpoints are outliers. In other words, if two endpoints of the line both are inliers, corresponding 2D-3D line matches of database images are reserved. And the 2D-3D candidate line match pairs between query images and 3D line map are acquired. 

The last step of line matching is according to reprojection error to get one-to-one 2D-3D line matches. 
 By performing two steps above, the one-to-many line correspondence between 2D line on the query image and 3D line in the world coordinate system is established. The reprojection error of the 3D line with query initial pose is computed which contains midpoint distance error and angle error that described in detail in Section \ref{pose_refiment}. 
 The high-quality line matches with the lowest distance and angle error are selected as the final line matches.

\begin{figure*}[htbp]
	\centering
	\includegraphics[width=0.8\textwidth]{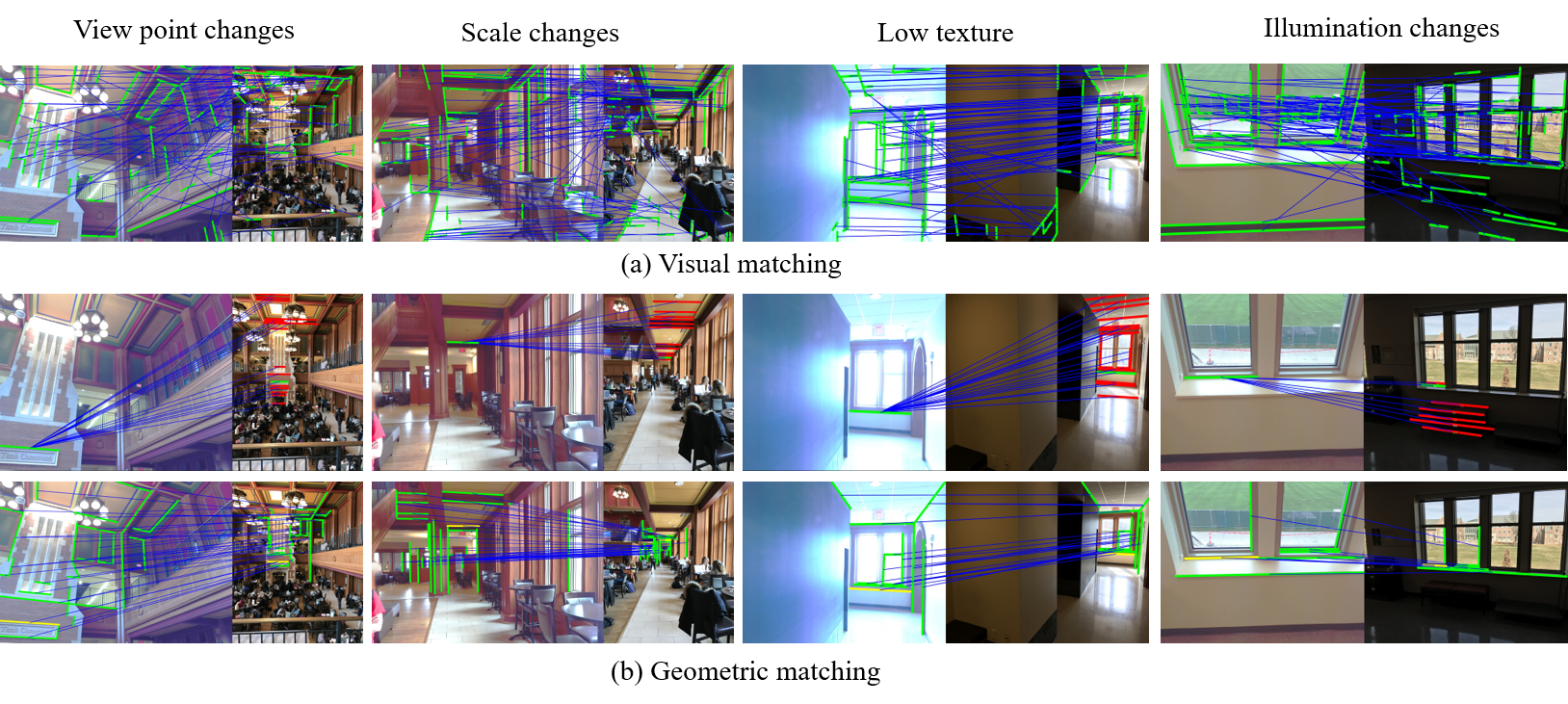}
	\caption{
		\textbf{Line matching result under different conditions.} (a) is visual matching result, using the state-of-the-art visual-based line matching method, SOLD2 \cite{pautrat2021sold2}. (b) is our geometric-based line matching result. the first row of (b) is the result based on epipolar constraint mentioned in Line3Dpp \cite{hofer2017efficient}. Since the matching pairs found by this method are one-to-many, here we only show the candidate matching of one line. The second row of (b) is the final result that obtained one-to-one line matching  pairs by reprojection filtering after epipolar constraint. From left to right, each column represents view point change, scale change, weak texture and illumination change respectively.}
	\label{fig:line_matching_result}
\end{figure*}

Fig. \ref{fig:line_matching_result} shows that our geometric-based method works well in some special scenarios, like view point changes, scale changes, weak texture and illumination changes, but visual-based method doesn't work for those scenarios.

\subsection{POSE REFINEMENT WITH POINT-LINE}
\label{pose_refiment}

For each query image, the initial pose is achieved by our previous work, RLOCS\cite{zhou2021retrieval}, a pure point-based localization. The 2D-3D line correspondences of the query image and the 3D line Map are computed based on the initial pose, which is described in detail in Section \ref{lines matchimg section}, denoted as ${Lcorr=\{(l_i,L_i)\}}$. The query SE(3) pose is denoted as ${T}$, and the reprojection of the 3D line ${L_i}$ on the query image plane is ${l_i^{\prime}}$. The pose ${T}$ is optimized by minimizing the midpoints distance error from ${l_i^{\prime}}$ to the corresponding 2D line ${l_i}$ and the angle error between them. The linear equations of ${l_i}$ and ${l_i^{\prime}}$ can be written as ${Ax+By+C=0}$ and ${A^{\prime}x^{\prime} + B^{\prime}y^{\prime} + C^{\prime}=0}$. Then the cost function can be formulated as follows:
\begin{equation} \label{eq: line_costfuction}
	{T^{*}=\mathop{\arg \min_{T}}\sum_{i=1}^{N}(\|d_i\|+\|{\theta}_i\|)}
\end{equation}
\begin{equation} \label{eq: distance_error}
	\|d_i\|=\|f(l_i) - f(h(K,T,L_i))\|
\end{equation}
\begin{equation} \label{eq: angle_error}
	\|{\theta}_i\|=\frac{{s}{|{A_i}{B_i^{\prime}}-{B_i}{A_i^{\prime}}|}}{\sqrt{{A_i}^{2}+{B_i}^{2}}\sqrt{{A_i^{\prime 2}}+{B_i^{\prime 2}}}}
\end{equation}
where ${N}$ represents the number of 2D-3D line correspondences, the function ${h(K,T,L_i)}$ equals ${l_i^{\prime}}$, which represents the reprojection of 3D line ${L_i}$ on the query image with pose ${T}$ and intrinsic ${K}$. There are two endpoints on ${l_i^{\prime}}$ and the function ${f({\cdot})}$ represents the mean value of them, known as the midpoint. The angle error ${\|{\theta}_i\|}$ is defined as the sine of the angle between ${l_i}$ and ${l_i^{\prime}}$. ${[A_i, B_i, C_i]}$ and ${[A_i^{\prime}, B_i^{\prime}, C_i^{\prime}]}$ are the coefficients of ${l_i}$ and ${l_i^{\prime}}$ respectively. However, the metric of distance error ${\|{d_i}\|}$ is in pixel. In order to keep the metric of angle error consistent with distance error, we apply a scale factor ${s}$ on the angle error. The scale is usually proportional to the length of ${l_i^{\prime}}$. 

The distance error ensures that there is enough overlap between the line pair, meanwhile, the angle error ensures that their directions are closed. Our combined reprojection distance is better than the traditional distance which count the only distance from projected endpoints to the corresponding 2D line, by avoiding the singular condition where two lines are close to collinear but non-overlapping.


Due to the input coarse pose resulting in the different perspective between the query image and database images, the only line features are not stable enough in some certain scenes. Therefore, we propose a point-line joint optimization method that combines the traditional point reprojection constraints with the above line constraints to make the optimization more robust. While obtaining the initial pose, the 2D-3D inlier point correspondences involved in pose solver are also preserved, denoted as ${Pcorr=\{(p_j, P_j)\}}$. Then the cost function evolves into the following form:
\begin{equation} \label{eq:costfunction}
	{T^{*}=\mathop{\arg \min_{T}}({\alpha}\sum_{i=1}^{N}(\|d_i\|+\|{\theta}_i\|)}+{\beta}\sum_{j=1}^{M}\|e_j\|)
\end{equation}
\begin{equation}
	\|e_j\|=\|p_j-h(K,T,P_j)\|
\end{equation}
where ${M}$ represents the number of 2D-3D point correspondences.
Similar to the formula \ref{eq: distance_error}, the function ${h(K,T,P_j)}$ represents the reprojection of 3D point ${P_j}$.
In addition, two different weights ${\alpha}$ and ${\beta}$ are applied to the lines and points reprojection error respectively to balance the weights of them. Usually, the ratio between ${\alpha}$ and ${\beta}$ is inversely proportional to the point and line correspondences numbers, and the sum of them equals 1 finally. The cost function is a nonlinear least squares problem and the Ceres-Slover\cite{ceres-solver} is employed to solve the optimized pose with the L-M algorithm and Huber loss function in our implementation.

\section{EXPERIMENTS}

In this section, we evaluate the performance of line detection and point-line pose optimization respectively. All experiments are conducted on one NVIDIA Tesla V100 GPU with the CUDA 11.0 and Intel(R) Xeon(R) Glod 6142 CPU @ 2.60GHz.

\subsection{Performance on Lines Detection}

Wireframe \cite{huang2018learning} and YorkUrban \cite{denis2008efficient} are typical line segment detection datasets. The following prevalent metrics are used in previous wireframe parser tasks \cite{von2008lsd,huang2018learning,xue2019learning,huang2020tp,xue2020holistically,gu2021towards}, and we evaluate our model with the structural Average Precision ($sAP$) \cite{zhou2019end,xu2021line}.

We compare VLSE with other line detection methods. The results and comparisons based on $sAP$ and $AP^H$ are summarized in Table \ref{tab:res1}. To make a fair comparison, we use the model with Hourglass as the backbone, length and horizontal angle as predicted map (Ours-HG-LA) as default architecture. By changing the line representation, as mentioned in \ref{line representation}, we predict the displacement distance in the x and y directions to form Ours-HG-Dxy, and the result both on the Wireframe and YorkUrban dataset improves significantly over Ours-HG-LA. Compared with other models, Ours-HG-Dxy model achieves competitive performance with HAWP, and outperforms L-CNN in terms of $sAP^{10}$ on Wireframe. To achieve a higher prediction accuracy, Ours-HG-HB alters the basic convolution block in HG with the Hybrid Block, 
and obtain the state-of-the-art results on Wireframe and YorkUrban which shows its robustness and generalizability.

\begin{table}[htbp]
	\caption{Quantitative comparisons between line detection methods.}
	\label{tab:res1}
	\renewcommand\arraystretch{1.3}
	\setlength{\tabcolsep}{1.5mm}{
		\begin{tabular}{lcccccc}
			\toprule
			\multicolumn{1}{c}{\multirow{2}{*}{Method}} & \multicolumn{3}{c}{\textit{Wireframe Dataset} \cite{huang2018learning}} & \multicolumn{3}{c}{\textit{YorkUrban Dataset} \cite{denis2008efficient}}                                                                 \\ \cline{2-7}
			                                            & $sAP^5$                                                                 & $sAP^{10}$                                                               & $sAP^{15}$    & $sAP^5$       & $sAP^{10}$    & $sAP^{15}$    \\ \hline
			LSD \cite{von2008lsd}                       & 6.7                                                                     & 8.8                                                                      & /             & 7.5           & 9.2           & /             \\
			DWP \cite{huang2018learning}                & 3.7                                                                     & 5.1                                                                      & 5.9           & 1.5           & 2.1           & 2.6           \\
			AFM \cite{xue2019learning}                  & 18.5                                                                    & 24.4                                                                     & 27.5          & 7.3           & 9.4           & 11.1          \\
			L-CNN \cite{zhou2019end}                    & 58.9                                                                    & 62.9                                                                     & 64.9          & 24.3          & 26.4          & 27.5          \\
			TP-LSD \cite{huang2020tp}                   & 57.5                                                                    & 60.6                                                                     & /             & 25.3          & 27.4          & /             \\
			HAWP \cite{xue2020holistically}             & \textbf{62.5}                                                           & 66.5                                                                     & 68.2          & 26.1          & 28.5          & 29.7          \\
			M-LSD \cite{gu2021towards}                  & 56.4                                                                    & 62.1                                                                     & /             & 24.6          & 27.3          & /             \\  \hline
			\textbf{Ours-HG-LA}                         & 60.5                                                                    & 65.4                                                                     & 67.1          & 24.6          & 26.9          & 28.2          \\
			\textbf{Ours-HG-Dxy}                        & 61.1                                                                    & 66.1                                                                     & 68.4          & 25.1          & 27.5          & 29.1          \\
			\textbf{Ours-HG-HB}                         & 62.1                                                                    & \textbf{67.3}                                                            & \textbf{69.6} & \textbf{26.6} & \textbf{29.3} & \textbf{30.9} \\
			\bottomrule
		\end{tabular}}
\end{table}

\begin{figure}[htbp]
	\centering
	\includegraphics[width=0.4\textwidth]{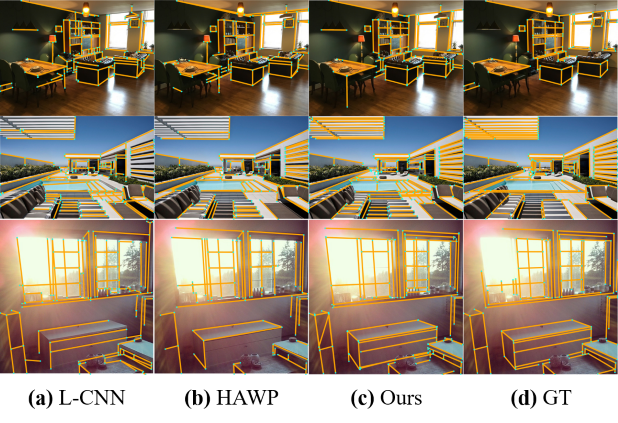}
	\caption{\textbf{Qualitative evaluation of line segment detection methods on Wireframe dataset.} From left to right, the columns show the results from (a) L-CNN \cite{zhou2019end}, (b) HAWP \cite{xue2020holistically}, (c) Ours, and (d) the ground truth.}
	\label{fig:line_viz}
\end{figure}

We also visualize the output of our VLSE and other methods L-CNN and HAWP in Fig. \ref{fig:line_viz}. Junctions and line segments are colored with cyan blue and orange, respectively. L-CNN, which relies on the junctions detection and might be prone to the missed junctions and nearby texture variation, captures many messy lines. HAWP gives a relatively precise detection result but seems to miss some key line segments.

\subsection{Performance on Localization}
\begin{figure}[htbp]
	\centering
	\includegraphics[width=0.4\textwidth]{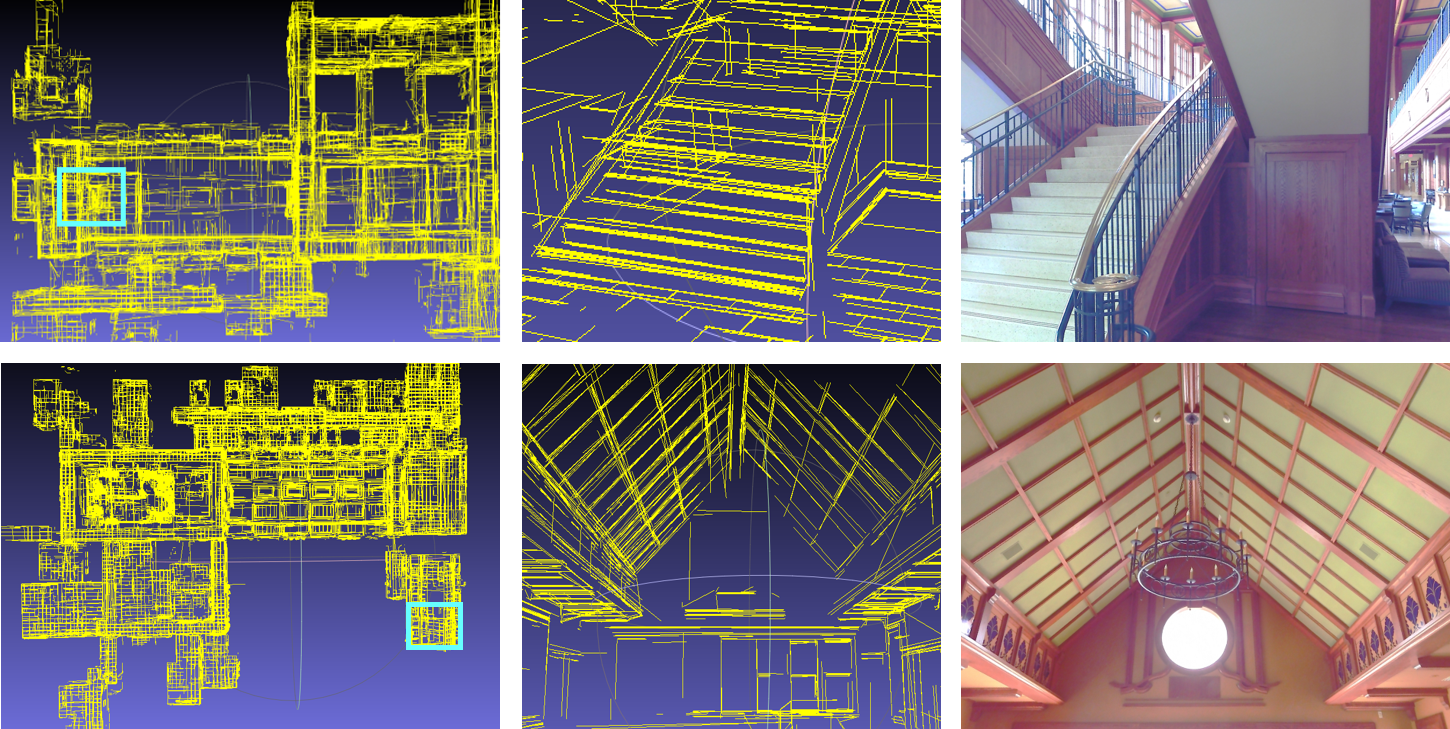}
	\caption{\textbf{Results of line mapping.} The first and second rows represent the result of the line map of InLoc \textit{duc1} and \textit{duc2} respectively. The first column represents a bird's-eye view, the second and the third column show the zoomed view of details and the actual scene.}
	\label{line_mapping_result}
\end{figure}

We validate our pose optimization method using InLoc datasets on the public Long-Term Visual Localization benchmark. 
InLoc is an indoor dataset that provides the coordinate value of each pixel in the world coordinates system. And it contains two scenarios, \textit{duc1} and \textit{duc2}. Their 3D line maps are reconstructed by projecting 2D lines into 3D space using known world coordinate values. The both 3D line maps are visualized respectively in Fig. \ref{line_mapping_result}.


Our previous work, RLOCS \cite{zhou2021retrieval}, provides the latest localization results as the initial poses of our optimization and the inliers involved in pose solver are also saved to join to point-line joint optimization. The localization performance is shown in the score on the online list, which is equal to the percentage of query images within three different thresholds of rotation and translation error. In order to verify the improvement of localization accuracy caused by line optimization and point-line joint optimization, which corresponds to Eq. \ref{eq: line_costfuction} and Eq. \ref{eq:costfunction}, we conduct the experiments on above two datasets under the different cost functions with the same initial poses. The results are shown in Table \ref{pose_optimization}.

\begin{table}[htbp]
	\centering
	\caption{point-line pose optimization on InLoc \textit{duc1} and \textit{duc2}}
	\label{pose_optimization}
	\renewcommand\arraystretch{1.3}
	\setlength{\tabcolsep}{1.5mm}{
		\begin{tabular}{ccc}
			\toprule
			\multirow{2}*{\textit{InLoc}}                 & \textit{duc1}                                                & \textit{duc2}                                  \\
			~                                             & \multicolumn{2}{c}{{(0.25m, 10°) / (0.5m, 10°) / (1m, 10°)}}                                                  \\
			\hline
			BaseLine                                      & 47.0 / 71.2 / 84.8                                           & 61.1 / 77.9 / 80.2                             \\
			Line Optimization                             & \textbf{51.5} / 71.7 / 84.3                                  & \textbf{61.8} / 78.6 / 80.9                    \\
			Point-Line Joint Optimization                 & 50.5 / \textbf{72.7} / \textbf{86.9}                         & \textbf{61.8} / \textbf{79.4} / \textbf{84.0}  \\
			\bottomrule
		\end{tabular}}

\end{table}


For \textit{duc1}, there is a significant improvement at accuracy ${(0.25m, 10^{\circ})}$, but a slight decrease at accuracy ${(1m, 10^{\circ})}$ after pure line optimization. For \textit{duc2}, the accuracy of each column is improved after pure line optimization. In addition, whether \textit{duc1} or \textit{duc2}, the result is further improved in different accuracy ranges after point-line joint optimization, especially at accuracy ${(1m, 10^{\circ})}$.

Moreover, we compare our point-line joint pose optimization method with various typical approaches on the Long-Term Visual Localization benchmark. The latest results of some outstanding approaches with the publication are captured from \url{visuallocalization.net/benchmark/} and shown in the Table \ref{benchmarch}. We found that our point-line joint optimization result named PtLine in benchmark list can not only improve the performance based on the baseline, but also can perform well compared to various excellent methods.

\begin{table}[htbp]
	\centering
	\caption{evaluation of excellent approaches on InLoc \textit{duc1} and \textit{duc2}}
	\label{benchmarch}
	\renewcommand\arraystretch{1.3}
	\setlength{\tabcolsep}{0.5mm}{
		\begin{tabular}{ccc}
			\toprule
			\multirow{2}*{\textit{InLoc}}                                           & \textit{duc1}                                                & \textit{duc2}                                  \\
			~                                                                       & \multicolumn{2}{c}{{(0.25m, 10°) / (0.5m, 10°) / (1m, 10°)}}                                                  \\
			\hline
			{KAPTURE-R2D2-FUSION\cite{humenberger2020robust}}                       & 41.4 / 60.1 / 73.7                                           & 47.3 / 67.2 / 73.3                             \\
			{DensePE+SNCNet+DensePV\cite{Rocco20}}                                  & 47.0 / 67.2 / 79.8                                           & 43.5 / 64.9 / 80.2                             \\

			{HLoc-SuperPoint+SuperGlue\cite{sarlin2020superglue}}                   & 49.0 / 68.7 / 80.8                                           & 53.4 / 77.1 / 82.4                             \\
			{SuperGlue+Patch2Pix\cite{sarlin2020superglue}\cite{zhou2021patch2pix}} & 50.0 / 68.2 / 81.8                                           & 57.3 / 77.9 / 80.2                             \\
			{LoFTR\cite{DBLP:journals/corr/abs-2104-00680}}                         & 47.5 / 72.2 / 84.8                                           & 54.2 / 74.8 / \textbf{85.5}                    \\
			{RLOCS-v3.0\cite{zhou2021retrieval}}                                    & 47.0 / 71.2 / 84.8                                           & 61.1 / 77.9 / 80.2                             \\
			\textbf{Ours (PtLine)}                                                  & \textbf{50.5} / \textbf{72.7} / \textbf{86.9}                & \textbf{61.8} / \textbf{79.4} / 84.0           \\
			\bottomrule
		\end{tabular}}

\end{table}

\section{CONCLUSIONS}

In conclusion, we propose a complete point-line optimization pipeline for pose refinement. We design a more accurate and stable line extractor for the following line matching and mapping process. And we adopt a geometric-based line matching method based on epipolar constraint with large line reprojection error filtering to get high-quality line matches. In addition, we establish a point-line joint cost function to optimize the query pose with an initial value obtained from the only-point localization.
Sufficient experiments on InLoc \textit{duc1} and \textit{duc2} are conducted to confirm the effectiveness and superiority of our proposed method. The pipeline is highly modular and extensible. In future, we look forward to further improvement of line extraction CNN by extending the training datasets into outdoor scenes, and escalating line matching and optimization strategy to reach the better localization precision.





\bibliographystyle{ieeetr}
\bibliography{root_bib}

\end{document}